%% file: emnlp2022.tex
\pdfoutput=1

\documentclass[11pt]{article}

\usepackage[]{EMNLP2022}

\usepackage{times}
\usepackage{latexsym}
\usepackage{cleveref}

\usepackage[T1]{fontenc}

\usepackage[utf8]{inputenc}

\usepackage{microtype}

\usepackage{adjustbox}
\usepackage{graphicx}
\usepackage{booktabs}
\usepackage{multirow}

\RequirePackage[inline,shortlabels]{enumitem}
\setlist{nosep}

\RequirePackage[textsize=footnotesize, color=yellow!20]{todonotes}

\usepackage{inconsolata}

\newcommand{\dataset}{{\scshape AraTweet}}
\newcommand{\transptc}{{\scshape TransPTC}}
\newcommand{\ctdtransptc}{{\scshape CtdTransPTC}}
\newcommand{\transptcplus}{{\scshape TransPTC+}}
\newcommand{\ctdtransptcplus}{{\scshape CtdTransPTC+}}

\def\ztitle{IITD at the WANLP 2022 Shared Task:\\ Multilingual Multi-Granularity Network for Propaganda Detection}
\title{\ztitle}

\author{
  Shubham Mittal\textsuperscript{1} \hspace{1em} \textbf{Preslav Nakov}\textsuperscript{2}\\
  \textsuperscript{1} Indian Institute of Technology Delhi \\
  \textsuperscript{2} Mohammed Bin Zayed University of Artificial Intelligence \\
  \texttt{shubhamiitd18@gmail.com, preslav.nakov@mbzuai.ac.ae}
}

\begin{document}
\maketitle

\begin{abstract}
\input{sections/abstract}
\end{abstract}

\section{Introduction}
\label{sec:intro}
\input{sections/intro}

\section{Data}
\label{sec:data}
\input{sections/data}

\section{System Description}
\label{sec:system}
\input{sections/system}

\section{Experiments}
\label{sec:exps}
\input{sections/experiments}

\section{Conclusion}
\label{sec:conclusion}
\input{sections/conclusion}

\bibliography{anthology,custom}
\bibliographystyle{acl_natbib}

\end{document}

%% file: sections/abstract.tex
We present our system for the two subtasks of the shared task on propaganda detection in Arabic, part of WANLP'2022.
Subtask~1 is a multi-label classification problem to find the propaganda techniques used in a given tweet.
Our system for this task uses XLM-R to predict probabilities for the target tweet to use each of the techniques.
In addition to finding the techniques, Subtask~2 further asks to identify the textual span for each instance of each technique that is present in the tweet; the task can be modeled as a sequence tagging problem.
We use a multi-granularity network with mBERT encoder for Subtask~2. Overall, our system ranks second for both subtasks (out of 14 and 3 participants, respectively).
Our empirical analysis show that it does not help to use a much larger English corpus annotated with propaganda techniques, regardless of whether used in English or after translation to Arabic.\footnote{The code is released at \href{https://github.com/sm354/mMGN.git}{github.com/sm354/mMGN}}

%% file: sections/intro.tex
\begin{table*}[htp!]
\centering
\small
\begin{tabular}{@{}lllllll@{}}
\toprule
\bf Propaganda Technique                                                    & \multicolumn{2}{c}{\bf train}                              & \multicolumn{2}{c}{\bf dev}                                & \multicolumn{2}{c}{\bf test}                               \\ \midrule
                                                    & \multicolumn{1}{c}{count} & \multicolumn{1}{c}{length} & \multicolumn{1}{c}{count} & \multicolumn{1}{c}{length} & \multicolumn{1}{c}{count} & \multicolumn{1}{c}{length} \\ \cmidrule(l){2-7} 
Appeal to authority                                 & 21          & 93.4 $\pm$ 43.9     & 8          & 94.8 $\pm$ 37.3          & 1          & 142.0 $\pm$ 0.0            \\
Appeal to fear/prejudice                            & 48          & 49.2 $\pm$ 29.0     & 11         & 54.9 $\pm$ 38.0          & 25         & 44.8 $\pm$ 27.9            \\
Black-and-white Fallacy/Dictatorship                & 2           & 60.5 $\pm$ 12.5     & 3          & 56.3 $\pm$ 20.4          & 7          & 49.6 $\pm$ 19.8            \\
Causal oversimplification                           & 4           & 80.0 $\pm$ 43.2     & 2          & 57.0 $\pm$ 18.0          & 4          & 57.3 $\pm$ 24.2            \\
Doubt                                               & 29          & 52.4 $\pm$ 34.6     & 3          & 61.0 $\pm$ 53.7          & 19         & 39.5 $\pm$ 21.3            \\
Exaggeration/Minimisation                           & 44          & 23.7 $\pm$ 28.4     & 26         & 14.3 $\pm$ 6.8           & 26         & 29.1 $\pm$ 16.9            \\
Flag-waving                                         & 5           & 57.6 $\pm$ 30.7     & 4          & 65.0 $\pm$ 19.6          & 9          & 60.1 $\pm$ 23.2            \\
Glittering generalities (virtue)                    & 25          & 81.4 $\pm$ 48.9     & 9          & 66.1 $\pm$ 17.2          & 1          & 104.0 $\pm$ 0.0            \\
Loaded language                                     & 446         & 9.70 $\pm$ 7.10     & 88         & 12.7 $\pm$ 13.2          & 326        & 7.20 $\pm$ 4.70            \\
Misrepresentation of someone's position             & 0           & N/A                 & 0          & N/A                      & 1          & 37.0 $\pm$ 0.0            \\
Name calling/Labeling                               & 244         & 13.8 $\pm$ 6.4      & 77         & 15.6 $\pm$ 8.4           & 163        & 14.1 $\pm$ 6.6            \\
Obfuscation, intentional vagueness, confusion       & 9           & 48.8 $\pm$ 28.1     & 4          & 34.0 $\pm$ 22.1          & 6          & 43.3 $\pm$ 23.6            \\
Presenting irrelevant data (red herring)            & 1           & 61.0 $\pm$ 0.0      & 0          & N/A                      & 0          & N/A            \\
Reductio ad hitlerum                                & 0           & N/A                 & 0          & N/A                      & 0          & N/A            \\
Repetition                                          & 9           & 12.8 $\pm$ 11.0     & 3          & 11.3 $\pm$ 4.1           & 3          & 35.3 $\pm$ 17.3            \\
Slogans                                             & 44          & 17.0 $\pm$ 6.6      & 2          & 26.5 $\pm$ 13.5          & 6          & 24.5 $\pm$ 11.7            \\
Smears                                              & 85          & 73.8 $\pm$ 34.9     & 27         & 88.8 $\pm$ 53.3          & 50         & 55.8 $\pm$ 22.0            \\
Thought-terminating cliché                          & 6           & 28.2 $\pm$ 17.5     & 2          & 21.0 $\pm$ 7.0           & 0          & N/A             \\
Whataboutism                                        & 3           & 47.7 $\pm$ 15.3     & 2          & 64.5 $\pm$ 20.5          & 0          & N/A             \\
Bandwagon                                           & 0           & N/A                 & 0          & N/A                      & 0          & N/A            \\ \midrule
no technique                                        & 95          & N/A                 & 15         & N/A                      & 44         & N/A            \\ \bottomrule
\end{tabular}
\caption{Instance count of propaganda techniques and their span length in characters (mean $\pm$ std-dev) in the \dataset{} partitions. N/A is for either \textit{no technique} or for those propaganda techniques having zero instances to compute mean/std-dev (such as \textit{Misrepresentation of Someone's Position}, \textit{Reductio ad hitlerum}, and \textit{Bandwagon}).}
\label{tab:labelstats}
\end{table*}

Propaganda is information deliberately designed to promote a particular point of view and to influence the opinions or the actions of individuals or groups.
With the rise of social media platforms, the circulation of propaganda is even more pronounced since it may be built upon a true fact, but exaggerated and biased to promote a particular viewpoint.
Various propaganda detection systems have been developed in recent years \cite{mgn2019, proppy1, proppy2,dimitrov-etal-2021-detecting, dimitrov-etal-2021-semeval}, but they all have been restricted to English due to the unavailability of labelled datasets (containing fine-grained annotations of textual spans) in other languages.
To bridge this gap, the WANLP'2022 shared task on propaganda detection in Arabic \cite{propaganda-detection:WANLP2022-overview} released a dataset of Arabic tweets (we will call it \dataset{}) that uses 20 propaganda techniques, thus enabling research beyond English.


There are two subtasks defined in this shared task for detecting the propaganda techniques used in a tweet: 
\begin{enumerate*}[(1)]
\item identify the techniques present in the given Arabic tweet, and
\item identify the span(s) of use of each technique along with the technique.
\end{enumerate*}
Subtask~1 can be viewed as a multi-label classification problem, where the tweet may contain any subset of the 20 propaganda techniques, even all or none of them.
Subtask~2 can be seen as a multi-label sequence tagging problem, where the system needs to predict the labels for each of the tokens.
Subtask~2 is more challenging than Subtask~1 due to the increased level of detail it asks for.

Our Subtask~1 system uses a multilingual pretrained language model, XLM-R \cite{xlmr} to estimate a Multinoulli distribution over the 20 propaganda techniques for a given Arabic tweet.
For Subtask~2, we use the multi-granularity network (MGN) from \citet{mgn2019}, but we replace the BERT encoder with mBERT \cite{bert}. We call our resulting system mMGN.
Our systems, which use only \dataset{} data, rank second for both subtasks.

We investigated cross-lingual propaganda detection by using the Propaganda Techniques Corpus (PTC) \cite{mgn2019}, which consists of annotated English news articles.
We trained mMGN on PTC and continued its training on \dataset{}.
Surprisingly, we found that continued training hurts the model by 10.2 F1 points absolute.
To alleviate the possibility of ineffective transfer from English in mBERT embeddings, we further translated the PTC to Arabic using Google Translate and we projected the span-labels using \texttt{awesome-align} \cite{awesomealign}.
Upon doing continued training with a subset of the translated data, having only sentences containing propaganda, we found that it does not help, but also does not hurt the model.
We believe that the domain difference between the two dataset is quite large, and thus there are no benefits in cross-lingual transfer.

%% file: sections/data.tex
The dataset released in this shared task, which we call \dataset{}, comprises Arabic tweets, most of which (but not all) contain some propaganda techniques.

\Cref{tab:labelstats} shows statistics about the propaganda technique in the partitions of \dataset{}.
Techniques such as \textit{Misrepresentation of Someone's Position (Straw Man)}, \textit{Presenting Irrelevant Data (Red Herring)}, \textit{Reductio ad hitlerum}, and \textit{Bandwagon} are rarely present in the dataset.
\textit{Loaded Language} is the most frequently present technique, whereas \textit{Appeal to Authority} has the longest span.
There are also tweets present that do not contain propaganda (e.g., 95 tweets in the training set).

\Cref{tab:dataset_stats} shows aggregated statistics about all propaganda techniques in the different partitions\footnote{The \emph{dev} partition in this work refers to the combination of dev and dev\_test released in the shared task.} of the dataset.

\begin{table}[h!]
\small
\centering
\begin{tabular}{llll}
\hline
                      & \bf train           & \bf dev             & \bf test           \\ \hline
\#examples            & 504             & 103             & 323            \\
\#spans               & 1025            & 271             & 647            \\
tweet len (t)           & 15.8$\pm$6.1    & 18.6$\pm$9.9    & 15.4$\pm$5.0     \\
tweet len (c)       & 112.6$\pm$39.2  & 123.4$\pm$58    & 117.4$\pm$30.6 \\ \hline
\end{tabular}
\caption{Statistics about the \dataset{}. Tweet len is the average length in \# tokens (t) and \# characters (c).}
\label{tab:dataset_stats}
\end{table}

%% file: sections/system.tex
\begin{figure*}[htp!]
\centering 
\includegraphics[scale=0.2]{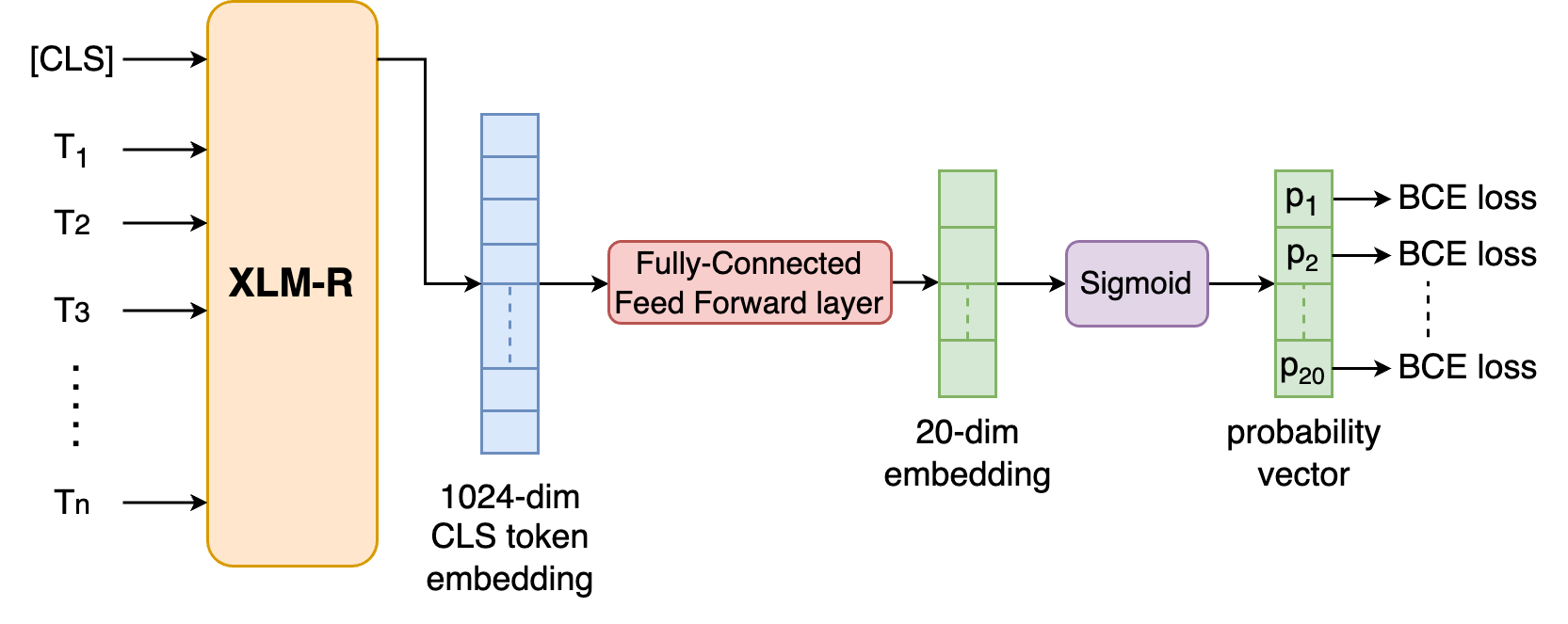} 
\caption{Our Subtask~1 system, which uses a pretrained XLM-R for multi-label classification. $T_1, T_2, T_3, \ldots, T_n$ are the tokens of the input tweet, and $p_i$ is the probability of the tweet using $i^{th}$ propaganda technique.}
\label{fig:mmgn}
\end{figure*}

\paragraph{Subtask~1} is a multi-label classification problem, where the model needs to find which of the 20 propaganda techniques (if any) are present in the input tweet.
Our system (shown in \Cref{fig:mmgn}) fine-tunes a multilingual pretrained language model, XLM-R, \cite{xlmr} for this subtask.

Given an Arabic tweet, we first tokenize it into word pieces $[T_1, T_2, \ldots, T_n]$ using the XLM-R tokenizer.
We the pass these pieces through XLM-R to obtain contextualised embeddings, from which we take the \textit{CLS} token embedding 
and we pass it through a single fully-connected linear layer to obtain a 20-dimensional embedding.
After passing it through a sigmoid non-linearity, we convert this embedding, representing logits, to probabilities $[p_1, p_2, ..., p_{20}]$, one for each propaganda technique.
Using a threshold of 0.5, our system assigns label $i$ if $p_i\geq0.5$.
When $p_i<0.5 ~ \forall ~ i$, the model predicts \textit{no technique} for the target tweet.

\paragraph{Subtask~2} is a multi-label sequence tagging problem, where we want to label the tokens of a given tweet with the propaganda techniques.
Since the (training) data contains tweets that do not contain propaganda (as discussed in \cref{sec:data}), we use the multi-granularity network (MGN) \cite{mgn2019} to develop our Subtask~2 system.

MGN uses BERT \cite{bert} and models the task as a single-label sequence tagging problem, where either one of 20 techniques or none of them is assigned to each token.
To improve the performance, it also adds a trainable \textit{gate} to lower the probabilities for all tokens if the sentence does not contain propaganda.\footnote{We refer the readers to \citet{mgn2019} for more detail.}

We replace BERT with mBERT in our MGN system, to obtain our multilingual multi-granularity network (mMGN) as our Subtask~2 system.
mMGN can work for Arabic and for all other languages that are supported by mBERT.

%% file: sections/experiments.tex
For evaluation, we use the official scorers that were released for the shared task.
The official evaluation measure for Subtask~1 is micro-F1. 
However, the scorer also reports macro-F1.
For Subtask~2, a modified micro-averaged F1 score is used, which gives credit to partial matches between the gold and the predicted spans.

We use the dev partition of \dataset{} to find the best model checkpoint and to report the scores on the finally released test set.
Our models are trained on a single V100 (32GB) GPU.

\paragraph{Subtask~1} We empirically compare different pre-trained language models (PLMs) as encoders for our Subtask~1 system and we report the scores in \Cref{tab:subtask1}.
With XLM-R encoder, our system achieves the best performance of 60.9 micro-F1.
The hyper-parameters of our Subtask~1 system include a maximum sequence length of 256, a batch size of 32, and 40 training epochs.
We use two different learning rates: 1e-5 for PLM and 3e-4 for the remaining trainable parameters.

\begin{table}[ht]
\small
\centering
\begin{tabular}{@{}lrr@{}}
\toprule
        &  macro-F1 &  micro-F1 \\ \midrule
mBERT \cite{bert}   &  8.1        & 54.3         \\
AraBERT \cite{arabert} & \textbf{18.7}     & 59.4     \\
XLM-R \cite{xlmr} & 18.3     & \textbf{60.9}     \\ \bottomrule
\end{tabular}
\caption{Performance(\%) of our Subtask~1 system with different multilingual pre-trained LMs.}
\label{tab:subtask1}
\end{table}

\paragraph{Subtask~2}
We train the multilingual Multi-Granularity Network (mMGN) model on \dataset{} with a batch size of 16, a learning rate of 3e-5 for PLM and 3e-4 for other trainable parameters, and 30 epochs.
This yields an F1 score of 35.5 on the test set, which is our best performance on this subtask.

\paragraph{Cross-lingual Propaganda Detection}
We ran several experiments using mMGN and the Propaganda Techniques Corpus (PTC), which is available in English \cite{mgn2019}, to study cross-lingual transfer between English and Arabic in Subtask~2.
In \begin{enumerate*}[(1)] \item \dataset{}, we train and test on \dataset{}, whereas in \item PTC, we train on PTC data and we test in a zero-shot manner on \dataset{}.
\item \transptc{} contains the translation of the PTC data from English to Arabic using Google Translate, followed by label projection using \texttt{awesome-align} \cite{awesomealign}.
Keeping only those translated sentences from \transptc{} that contain propaganda gives \item \transptcplus{}.
\item \ctdtransptc{} and \item \ctdtransptcplus{} take the trained model from \transptc{} and \transptcplus{}, respectively, and train it further on \dataset{}.
\end{enumerate*}

The performance across all settings is reported in \Cref{tab:subtask2}. We can see that \transptc{} is better than PTC by 0.6 F1 points, which suggests that the model learns better with the Arabic PTC.

\begin{table}[tbh]
\small
\centering
\begin{tabular}{@{}lrrr@{}}
\toprule
                     & Precision & Recall & F1    \\ \midrule
\dataset{}           & 35.5      & 25.7   & \textbf{29.8}  \\
PTC                  & 53.1      & 1.4    & 2.8   \\
\transptc{}          & 30        & 1.8    & 3.4   \\
\transptcplus{}      & 34.2      & 10.6   & 16.1  \\
\ctdtransptc{}       & 21        & 18.4   & 19.6  \\
\ctdtransptcplus{}   & 30.6      & 28.0   & 29.2  \\ \bottomrule
\end{tabular}
\caption{Performance(\%) of mMGN (on dev\_test) using different training methodologies.}
\label{tab:subtask2}
\end{table}

The 1.8 recall of \transptc{} is quite low, which could be due to the high proportion of propaganda-free sentences in PTC, which makes the model reluctant to propose propaganda techniques.
When training only on propaganda-containing translated sentences from PTC, \transptcplus{} improves over \transptc{} on recall and also on precision, resulting in a gain of 12.7 F1 points absolute.
Continued training on \dataset{}, \ctdtransptc{} and \ctdtransptcplus{} yields sizable gains over the PTC-trained models \transptc{} and \transptcplus{}.
However, \ctdtransptcplus{} is worse than \dataset{} by 0.6 F1 points absolute, indicating that cross-lingual transfer is not helping, but also not significantly hurting the performance.

We posit that the large domain difference between the PTC and the \dataset{} datasets may be the reason for ineffective cross-lingual transfer.
PTC contains news articles whereas \dataset{} contains tweets, which causes linguistic differences in the text such as the presence of URLs, emojis, or slang in the tweets.
Tweets are also often shorter due to text length limit in Twitter, which may also confuse the model between the two datasets.

%% file: sections/conclusion.tex
We described our systems for the two subtasks of the WANLP 2022 shared task on propaganda detection in Arabic. 
For Subtask~1, we used XLM-R to estimate a Multinoulli distribution over the 20 propaganda techniques for multi-label classification.
For Subtask~2, we used a multi-granularity network with mBERT, addressing the subtask as a sequence tagging problem. The official evaluation results put our systems as second on both subtasks, out of 14 and of 3 participants, respectively.
We further described a number of experiments, which suggest various research challenges for future work, such as how to effectively use data from different domains, and how to learn language-agnostic embeddings for propaganda detection.